\documentclass[letterpaper, 10 pt, conference]{packages/ieeeconf}
\IEEEoverridecommandlockouts

\usepackage{packages/algorithm}
\usepackage{packages/algorithmic}

\usepackage[compress]{cite}
\makeatletter
\let\NAT@parse\undefined
\makeatother
\usepackage[colorlinks=true]{hyperref}

\usepackage[table]{xcolor} 
\usepackage[section]{placeins} 

\usepackage{microtype} 
\usepackage{graphics} 
\usepackage{amsmath} 
\usepackage{mathtools} 
\usepackage{todonotes}
\usepackage{amssymb}  
\usepackage{makeidx} 
\usepackage{graphicx} 
\graphicspath{{figures/}}
\usepackage{multicol} 
\usepackage[bottom]{footmisc} 
\usepackage[utf8]{inputenc} 
\usepackage{multirow}
\usepackage{booktabs} 
\usepackage{adjustbox}
\usepackage{caption}
\usepackage{makecell} 
\usepackage{subcaption} 
\usepackage{tikz}
\usetikzlibrary{positioning}
\usetikzlibrary{shapes,snakes}
\usepackage{bm} 
\usepackage{float} 
\usepackage{color} 
\usepackage{empheq} 
\usepackage{threeparttable} 
\usepackage{xcolor}
\usepackage{xurl} 
\usepackage{tcolorbox}

\DeclareMathSymbol{\shortminus}{\mathbin}{AMSa}{"39}
\usepackage{paralist}

\newcommand\gbf[1]{\textbf{\color{green!80!black}{#1}}} 
\newcommand\gs[1]{\textbf{\textcolor{blue}{#1}}} 

\usepackage{mathtools}
\DeclarePairedDelimiterX{\norm}[1]{\lVert}{\rVert}{#1}

\allowdisplaybreaks
\begin{document}

\title{\LARGE \bf
Learning IMU Bias with Diffusion Model
}

\author{Shenghao Zhou$^{1}$, Saimouli Katragadda$^{1}$, Guoquan Huang$^{1}$
\thanks{This work was partially supported by 
the University of Delaware (UD) College of Engineering, 
NSF (SCH-2014264, IIS-2410019), 
Google ARCore,
and Meta Reality Labs.
}
\thanks{$^{1}$The authors are with the Robot Perception and Navigation Group (RPNG), University of Delaware, Newark, DE 19716, USA. 
Email: {\tt\small \{shzhou, saimouli, ghuang\}@udel.edu}}%
}

\maketitle
\pagestyle{empty}


\begin{abstract}
Motion sensing and tracking with IMU data is essential for spatial intelligence,
which however is challenging due to the presence of time-varying stochastic bias.
IMU bias is affected by various factors such as temperature and vibration, making it highly complex and difficult to model analytically.
%
Recent data-driven approaches using deep learning have shown promise in predicting bias from IMU readings.
However, these methods often treat the task as a regression problem, overlooking the stochatic nature of bias.
In contrast, we model bias, conditioned on IMU readings, as a probabilistic distribution and design a conditional diffusion model to approximate this distribution.
Through this approach, we achieve improved performance and make predictions that align more closely with the known behavior of bias.


\end{abstract}

\section{Introduction}


3D motion tracking is essential to endow mobile devices and autonomous vehicles with spatial intelligence. 
Due to the recent advancements in  MEMS sensing technology, 6-axis IMUs measuring angular velocity and linear acceleration have become ubiquitous and made it possible to estimate 3D motion for sensor platforms at edge with compact size, minimal weight, low power consumption and cost (SWaP-C).
However,  naive integration of IMU measurements to offer 3D odometry (i.e., acceleration, rotation and velocity) or dead reckoning -- without aiding sources such as GPS and vision -- often is not reliable and diverges in a very short period of time.
Better solutions of {\em inertial-only odometry (IOO)} than naive inertial integration are desperately needed in practice.
For example, consider hand tracking in mobile AR/VR applications, highly dynamic hands can easily move out of the tracking camera's field of view (FOV), leaving only IMU data available to keep motion tracking alive.

If IMU measurements were clean and noise free, then naive inertial integration would solve the IOO problem.
The reality is much bitter, primarily due to the time-varying stochastic biases that significantly corrupt the inertial signals. 
As such, in order to find a better IOO solution, it is almost inevitable to better find IMU bias,
which is precisely what this paper seeks to address.
%
%
%
%
%
%
IMU bias represents an offset of the output from the input value and encompasses many different types of bias parameters such as in-run bias stability, turn-on bias repeatability, and bias over temperature.
Many unforeseeable factors such as temperature and vibrations can affect the IMU bias, which makes it impossible to correctly model it~\cite{titterton2004strapdown}, 
although there are simplified but useful models such as random walk widely used in practice~\cite{furgale2013kalibrIROS,Geneva2020ICRA}.
%

%

With the emerging of deep learning, there are attempts to model IMU bias in a data-driven manner with neural networks~\cite{Cohen2024}. 
These approaches have demonstrated the possibility of regressing bias from IMU readings and subsequently integrating the IMU data to estimate motion with reasonable accuracy over  short periods.
%
%
In particular, one may use a differentiable integration module to integrate IMU readings with the predicted bias removed, and compare the result motion with the ground truth~\cite{Zhang2021ICRA, Qiu2024Arxiv}.
However, it cannot guarantee the predicted correction to IMU reading is the actual bias. %
This is because there exists other correction to the IMU reading that can achieve the same or even better integrated motion result, but very different from the true bias. %
When the supervision is provided indirectly through the integrated motion, the network can learn to make these spurious
predictions instead of the real bias. This may not generalize to new data, because the learned correction is not an intrinsic property of IMU, as bias does. 
Alternatively, one can directly use ground truth bias for supervision~\cite{Buchanan2023RAL}. %
This method currently only shows to work when integrated with camera in an optimization based VINS system. As we show in the experiment, the performance of this method is inferior compared with indirectly supervised methods. 
%
%
Both approaches assume a single true bias value for a given IMU reading, framing the problem as a regression task. 

In this paper, we  propose to model the IMU bias naturally as a probability distribution conditioned on the inertial reading, instead of a fixed value. 
%
This formulation, combined with direct supervision, allows for more accurate and faithful bias prediction.
To model this complex distribution, we leverage diffusion model, which has shown promising results in capturing distributions with high uncertainity in tasks such as action planning\cite{chi2024diffusionpolicyIJRR} and human trajectory prediction \cite{gu2022MID_CVPR}.
%
In particular, we design a conditional diffusion model that takes feature extracted from the IMU reading as an additional condition code to approximate the underlying IMU-conditioned bias distribution.
%
The IOO with the proposed diffusion model is shown to outperform the regression-based approaches (with both direct and indirect supervision). 
Additionally, our predicted bias closely resembles to the ground truth in terms of magnitude and variation patterns, showing superior accuracy and generalization.

In summary, the main contributions of this paper include: 
\begin{itemize}
    \item We, for the first time, design a lightweight diffusion model to learn IMU bias for IOO in a data-driven manner, by naturally modeling bias as a probability distribution conditioned on inertial measurements.
    %
    %

    \item We experimentally validate that the proposed diffusion model achieves more accurate bias prediction,
     confirming that our probabilistic modeling approach is effective, outperforming regression-based methods, both with direct and indirect supervision.
    %

\end{itemize}

\section{Related Work}


Many IOO methods exist and can be categorized into model-free and model-based approaches, depending on whether or not the IMU bias is explicitly modeled.

\subsection{Model-free Method}

Early work explores to leverage motion pattern, with primary applications in  Pedestrian Dead Reckoning (PDR) scenarios. Heuristic algorithms such as step counting algorithm with step length estimation \cite{Brajdic2013ACM}  and  stationary period detection with zero velocity update (ZUPT) \cite{Foxlin2005CGA} are explored. A system combine multiple heuristic algorithms working on mobile phone is presented in \cite{solinInertialOdometryHandheld2018}. 
In recent years, there is attempt to use deep learning neural network, to learn to regress the motion from IMU reading end-to-end \cite{yan2018ridi_ECCV}, \cite{chen2018ionet_AAAI}, \cite{Sun2021AAAI}, \cite{Herath2020ICRA}, \cite{Liu2020RAL}, \cite{herath2022nill_CVPR}, \cite{Cioffi2023learned}. These methods show promising results on PDR scenarios, suppressing the classical method. Positional displacement and velocity are explored as the target for network prediction. 
Some work leverages the equivariance in the IMU reading, as a way to enable self-supervised learning \cite{cao2022rio} or boost the performance \cite{jayanth2024eqnio}, further pushing the limit of this method. 
However, 
these methods still implicitly rely on motion pattern. Essentially, these methods use deep learning to capture motion pattern in a data-driven fashion. 
Noticeably, \cite{Cioffi2023learned} shows such end-to-end learning can work in drone-racing scenario, though it only works when training and testing is on the same trajectory. In this case, the high-speed drone motion for a particular trajectory becomes a complex motion pattern.  This shows deep learning can learn non-trivial motion pattern. Yet, it still can't break the theoretical limitation of the reliance on the patterned motion.  
In this work, we consider general scenario without patterned motion assumption. In this scenario, model-free method shows inferior performance because it struggles to find motion pattern.

\subsection{Model-based Method}
Model-based method aims to estimate the bias from IMU readings, then remove the bias, and use integration to get motion estimation. 
Early analysis of IMU bias shows  many factors such as temperature, vibration and impacts, all affect IMU bias \cite{titterton2004strapdown}.
However, the compound effect is hard to model with analytical model. 
Popular in the system with IMU and other sensors, random walk model \cite{trawny2005indirect} is a simplified choice for bias modeling. It models the bias evolution as a Brownian noise process. 
However, such model has limited accuracy, and it can't be used without other sensors. 
Also, it typically requires collecting long period stationary IMU readings for offline calibration to get model parameters. 

Recent deep learning methods offer new way to model bias. Since the end goal is to remove the bias, this approach is also referred as denoising approach. Since bias is not directly available as data, some approaches use indirect supervision from integrated motion, leveraging a differentiable integration process. 
The first work \cite{Brossard2020RAL} estimates gyro bias only, with integrated rotation as training data. \cite{Zhang2021ICRA} proposes to use supervision from integrated pre-integration terms to regress bias. 
Recent work \cite{Qiu2024Arxiv} uses integrated motion to regress both bias and its uncertainty. It achieves state of the art result on a few datasets. 
However, indirect supervision has a misalignment between their training target and the network output. Since multiple IMU readings can produce the same integrated result, supervising with integrated result can't guarantee the network can learn the actual bias instead of predicting other signals. Our experiment shows methods trained with indirect supervision will make spurious prediction that is very different from actual bias. This may hurt the generalization ability, since other signals might not generalize to new data even for the same IMU.

Close to our work, \cite{Buchanan2023RAL} proposes to use direct supervision from bias for training. However, it only demonstrates the performance when fusing the bias prediction with vision in a joint factor graph system. As our experiment shows, such direct supervision under regression setting will have limited accuracy. 
Our method follow the deep learning approach 
for bias modeling using direct bias supervision. However, different from all the work mentioned above, we deviates from the regression formulation, and treats the bias given IMU reading as a conditional probability distribution.

\section{Inertial-Only Odometry}
\label{problem_formulation}

While inertial navigation systems (INS) aided by different exteroceptive sensors (such as vision and GPS) have been widely studied in the literature (e.g., see \cite{Huang2019ICRA}), IOO requires further investigation as aiding sensors can easily degrade or fail in practice. 
In this section, we will revisit the IOO problem from an INS perspective 
while focusing on the bias modeling challenges.

\subsection{Inertial Navigation}

IOO shares the same IMU kinematics as INS to estimate motion (i.e., position, rotation and velocity) 
using IMU (accelerometer and gyroscope) measurements.
Each accelerometer measures proper acceleration on only one axis, and are therefore usually found in groups of three orthogonal devices on a single low cost MEMS chip.
However, low-cost accelerometer measurements are far from ideal and are corrupted by noise and bias:
\begin{align}\label{eq:acc-model1}
\mathbf a_m(t) &= \mathbf C(^I_G\bar{\mathbf q}(t))  \left( {^G\mathbf a}(t) - {^G\mathbf g}  \right) + \mathbf b_a(t) + \mathbf n_a(t)
\end{align}
where 
$^I_G\bar{\mathbf q}$ is the unit quaternion 
that represents the rotation from the global frame of reference $\{G\}$ to the IMU frame $\{I\}$ 
(i.e., corresponding to the rotation matrix $\mathbf C(^I_G\bar{\mathbf q}) $),
${^G\mathbf a}$ is the true acceleration of the IMU in the global frame $\{G\}$,
${^G\mathbf g}$ is the gravitational acceleration expressed in $\{G\}$, 
and $\mathbf n_a\sim \mathcal N(\mathbf 0, \mathbf N_a)$ is zero-mean, white Gaussian noise,
and $\mathbf b_a$ is the bias changing over time.
%
Like the accelerometer, gyroscope measures angular velocity of the sensor and suffers from noise and bias, and sometimes, misalignment and scale errors.
Moreover, gyroscope measurements are also influenced by acceleration (i.e.  g-sensitivity),
whose magnitude  is negligible if it is within the range of the additive white noise $\mathbf n_g$, 
while in some low-cost MEMS hardware, it can be more significant:
\begin{align} \label{eq:gyro-model3}
\bm\omega_m(t) &= \mathbf T_g {^I\bm\omega}(t) + \mathbf T_s {^I\mathbf a}+ \mathbf b_g(t) + \mathbf n_g(t) 
\end{align}
where 
$\mathbf T_g$  is the shape matrix causing both misalignment and scale errors in the gyro measurements,
$\mathbf T_s$ is the g-sensitivity coefficient,
$\mathbf n_g \sim \mathcal N (\mathbf 0, \mathbf N_g)$ is zero-mean white Gaussian noise,
and the bias $\mathbf b_g$ is time-varying and random.
%

The INS kinematic model is given by~\cite{trawny2005indirect}:
\begin{align}
^I_G\dot{\bar{\mathbf q}}(t) &= \frac{1}{2} \bm\Omega\left(^I\bm\omega(t)\right) {^I_G\bar{\mathbf q}}(t)\\
^G\dot{\mathbf p}(t) &= {^G\mathbf v(t)} \\
^G\dot{\mathbf v}(t) &= {^G\mathbf a}(t) 
\label{eq:imu-motion}
\end{align}
where 
$^I\bm\omega = \begin{bmatrix} \omega_1 & \omega_2 & \omega_3 \end{bmatrix}^T$ is the true rotational velocity of the IMU,  and $\bm\Omega(\bm\omega)$ is defined by:
\begin{align*}
\bm\Omega(\bm\omega) = \begin{bmatrix}
-\lfloor \bm\omega \times \rfloor & \bm\omega \\
-\bm\omega^T & 0 \end{bmatrix}~,~~
\lfloor \bm\omega \times \rfloor = \begin{bmatrix}
0 & -\omega_3 & \omega_2 \\
\omega_3 & 0 & -\omega_1 \\
-\omega_2 & \omega_1 & 0
\end{bmatrix}
\end{align*}
Using the IMU measurements and assuming {\em known} bias models (e.g., random walk), 
3D motion estimates can be obtained by integrating the above continuous-time kinematics.
Clearly, the quality of IMU data (affected by noise and bias) determines the motion accuracy.

Because of the (aided) INS observability properties~\cite{Yang2019TRO},
any method that tries to bypass bias modeling and directly predict
global position or velocity has the fundamental limitation on the target motion pattern. 
As we focus on general scenario without prior  motion pattern assumption, we limit to estimate motion increment (i.e., odometry), while only assuming known initials if absolute motion is needed. 

\subsection{Modeling Bias}


As evident, it is critical to find biases for IOO from IMU measurements in order to be able to perform accurate inertial integration to estimate motion:
\begin{equation}
\label{eq:f-pi}
\begin{bmatrix}
\mathbf{b_g}(t) \\ 
\mathbf{b_a}(t) 
\end{bmatrix} = f_{\pi}(\boldsymbol{\omega_m}(t), \mathbf{a_m}(t))
\end{equation}
where $f_\pi$ is some estimator.
However, finding such estimator is non-trivial because the bias is not deterministic. As an IMU is a physical electronic sensor,  factors such as temperature, impacts, vibration, and quantization noise all affect it \cite{titterton2004strapdown}. 
These compound effects are complex and difficult to model.
Moreover, many of these factors are time-varying, giving the bias a stochastic nature.
Not only does the bias change as the IMU operates, but also after the power cycles further complicating model development.
As such, it is almost impossible to analytically model the IMU bias.


While building an exact model is challenging, analysis on bias as a black-box signal using Power Spectral Density (PSD) \cite{witt1997_PSD_TIM} and Allan variance analysis \cite{allan1987_allanvariance_TIM} reveal certain bias characteristic, such as angle/velocity random walk, bias instability and rate ramp. These characteristics become standard in the industry for IMU sensors   
~\cite{ieee_gyro_standard_2004, ieee_acc_standard_2019}.
However, utilizing all these characteristics to build an estimator is difficult because some of them, like bias instability are defined only in frequency domain, without state-space equivalent.

As approximation,  in practice, a simplified model leveraging rate random walk is commonly used as the bias model ~\cite{furgale2013kalibrIROS,Geneva2020ICRA}. 
Specifically, it assumes the bias dynamic model as:
\begin{equation}
\begin{bmatrix}
\dot{\mathbf{b}}_g(t) \\ 
\dot{\mathbf{b}}_a(t) 
\end{bmatrix} = 
\begin{bmatrix}
\eta_g(t) \\ 
\eta_a(t) 
\end{bmatrix}
\end{equation}
To fit parameters $\eta_g, \eta_a$, a common approach is to collect a sequence of stationary IMU readings and fit them with using Allan variance analysis, e.g., as demonstrated in Kalibr \cite{furgale2013kalibrIROS}:
\begin{equation}
\begin{bmatrix}
\eta_g(t) \\ 
\eta_a(t) 
\end{bmatrix} = \textrm{calibration} (
\boldsymbol{\omega}_\textrm{m\_static}, \mathbf{a}_\textrm{m\_static})
\end{equation}
The initial values $\mathbf{b_g}(0), \mathbf{b_a}(0)$ require extra heuristics to estimate, such as taking the average of stationary IMU reading and subtract.
Although this simple model captures the slow variations in bias, its accuracy is limited and typically requires external sensors to aid inertial navigation.
Additionally, the process involves two steps: first, estimating dynamic parameters using specific IMU readings, and second estimating bias based on the dynamic model.
The first step calibration is not only time-consuming but also restrictive, as it demands an extended period of stationary IMU readings.

\section{Learning Bias for IOO}

In this section, we thus design a deep neural network to represent the modeling function $f_\pi$~\eqref{eq:f-pi},
which can be trained end-to-end to predict the IMU bias.
This is in contrast to the classical random walk model, which uses a hand-craft two-step pipeline and assumes a long period static IMU reading available. 
These models are shown to be able to generalize to unseen readings with good accuracy. 
The success motivates us to take the approach of deep learning based bias modeling. 
\begin{equation}
\begin{bmatrix}
\mathbf{b_g(t)} \\ 
\mathbf{b_a(t)} 
\end{bmatrix} = \textrm{network}(\boldsymbol{\omega_m}, \mathbf{a_m})
\end{equation}
However, different from the literature, we do not treat it as a regression problem, assuming $\mathbf{b_g}, \mathbf{b_a}$ are fixed value. %
Instead, we model them as probability distribution, as
$p(\mathbf{b_g}, \mathbf{b_a} | \boldsymbol{\omega}, \mathbf{a})$. %
This probability distribution can be very complex, thus deep learning model is a good fit to estimate them.

\subsection{Diffusion Model}

Diffusion models \cite{Ho2020NeurIPS} are generative models that aims to represent data $\mathbf{x}_0$ using a series of latent codes $\mathbf{x}_1, \dots, \mathbf{x}_T$ through a forward and reverse diffusion process.
The forward process gradually adds noise to the data, encoding it into a structured latent space, while the reverse process decodes the latent code back into the original data.
%
%
Once trained, the model allows us to sample a latent code $\mathbf{x}_T$ from a simple distribution and generate the corresponding data $\mathbf{x}_0$ by running the reverse diffusion process.
The key strength of the diffusion model lies in its ability to model highly complex distribution $\mathbf{x}_0 \sim q(x)$, by leveraging the multiple latent representations between $\mathbf{x}_1$ and $\mathbf{x}_T$.  That is why we want to use diffusion model to learn conditional bias distribution. 
The latent space is structured in such a way that $\mathbf{x}_T$ follows a simple Gaussian distribution, making sampling straightforward.

The encoding process between two latent codes $\mathbf{x}_{t-1}, \mathbf{x}_t$ is performed by adding Gaussian noise.
\begin{align}
\mathbf{x}_t = \sqrt{1-\beta_t} \mathbf{x}_{t-1} + \sqrt{\beta_t} \epsilon_{t-1}, \epsilon_{t-1} \sim \mathcal{N}(0,1)
\end{align}
where $\beta_t$ is the hyperparameter that controls the amount of noise added at each step $t$, and $T$ is the total number of diffusion steps. As $t$ increases, the latent variable $\mathbf{x}_t$ transitions towards pure Gaussian noise.

At the core of the diffusion model is the denoiser network, which is described in Sec. \ref{sec:system}, aiming to estimate the noise added at each step in the forward process.
Given corrupted data $\mathbf{\mathbf{x}_t}$ and the step $t$, the network predicts the noise $\epsilon_{t-1}$ added at the previous step: 
\[
\hat{\epsilon}_{t-1} = \epsilon_\theta(\mathbf{x}_t, t)
\]
The denoiser network is trained using the Mean Squared Error (MSE) loss on the noise: 
\begin{equation}
\lVert\epsilon_{t-1} - \hat{\epsilon}_{t-1}\rVert_2
\label{equ:diffusion_loss}
\end{equation}
This simple training loss function is equivalent to minimizing the evidence lower bound (ELBO) from variational inference perspective, which allows the model to approxiamte the  underlying distribition of data $\mathbf{x}_0$.
To generate a sample from the diffusion model, we first sample a latent code $\mathbf{x}_T$, and decode it back to original data $\mathbf{x}_0$. 
%
$\mathbf{x}_T$ follows a Gaussian distribution, as conceptually it is the result of adding Gaussian noise for T steps in the forward process. The sampling process begins with:  

\begin{equation}
\mathbf{x}_T \sim \mathcal{N} (0, I)
\end{equation}
Next, we use denoiser network to iteratively decode $\mathbf{x}_t$ back to $\mathbf{x}_{t-1}$ as follows:
\begin{equation}
\label{equ:denoise}
\mathbf{x}_{t-1} = \frac{1}{\sqrt{1-\beta_t}} (\mathbf{x}_t - \gamma_t \epsilon_\theta(\mathbf{x}_t, t)) + \sigma_t z, z \sim \mathcal{N}(0, I) 
\end{equation}
where $\beta_t, \sigma_t, \gamma_t$ are fixed value. The parameter $\beta_t$ is the same noise variable used in the forward process, while both $\beta_t, \sigma_t$ are  hyperparameters of the diffusion model, controlling the noise schedule. %
The parameter $\gamma_t$ is a fixed function of $\beta_t$.

%


%
In our bias modeling, $\mathbf{x}_0$ corresponds to the original bias $(\mathbf{b_g}, \mathbf{b_a})$. The bias is the only required training data. 

As we want to model the conditional probability distribution of bias given IMU readings, we introduce an additional feature vector $\mathbf{c}$ extracted from the IMU readings.
This feature  $\mathbf{c}$ serves as conditional code to the denoiser network at each step $t$ of the denoising process, so that we can model the conditional distribution:
\begin{equation}
\hat{\epsilon}_{t-1} = \epsilon_\theta(\mathbf{x}_t, t, \mathbf{c})
\label{equ:conditonal_network}
\end{equation}
The training and sampling steps  remain the same.
%
%

%


\begin{figure}[t]

\includegraphics[width=0.5\textwidth]
{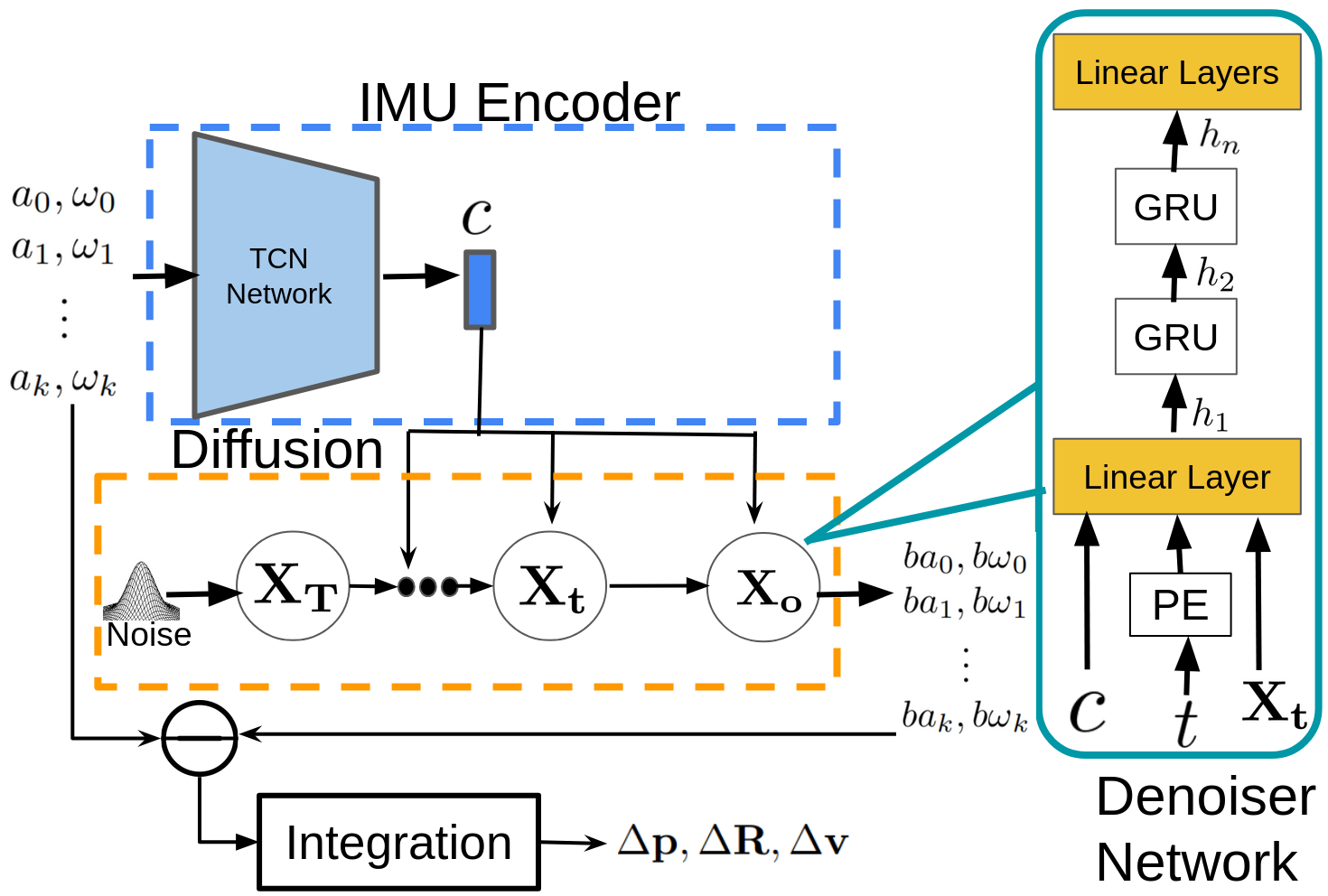}
\caption{System overview: our model consists of IMU encoder and denoiser network of the diffusion model.  Conditional code $\mathbf{c}$ extracted by IMU encoder from IMU readings is passed to the denoiser network, to generates the bias with multiple diffusion steps. Bias is used to correct the IMU readings for integration, to get the motion estimation.}
\label{fig:overview}
\end{figure}

\subsection{Model Design}
\label{sec:system}
As shown in Fig. \ref{fig:overview}, we design two models to implement $\epsilon_\theta$ in equation \ref{equ:conditonal_network}: the IMU encoder and the denoiser network of the diffusion model. The IMU encoder extracts feature code $\mathbf{c}$ and pass it to the denoiser network, as the implementation of $\epsilon_\theta(\mathbf{x}_t, t, \mathbf{c})$. 
%
%

We choose Temporal Convolutional Network (TCN) as the IMU encoder, because 
%
it effectively captures the temporal relation in sequential data. It is easy to train and deploy because it mainly consists of convolutional layers \cite{Bai2018_TCN_Arxiv}. Previous deep learning based IOO work ~\cite{Herath2020ICRA, Liu2020RAL} shows it can extract useful information in IMU reading sequence. 
%
%

%
The second component is the denoiser network for the diffusion model. It needs to fuse the conditional code $\mathbf{c}$ from IMU encoder with diffusion model latent code $x_t$ and step number $t$, and then process the fused code with its backbone to make prediction, and optimize for the loss function in equation \ref{equ:diffusion_loss}. The internal structure is illustrated on the right of Fig.  \ref{fig:overview}. 
%
%
%

The fusion is done with one linear layer, as a simple design. 
Since in diffusion models, each denoising step corresponds to a specific noise level, the timestep information is critical.
Thus, we add sinusoidal positional embedding to the step number $t$ to provide a smooth, continuous representation of time, inspired by the design of transformer \cite{vaswani2017attention}.

Deviating from U-Net \cite{Ronneberger2015Springer}, a popular design for diffusion models, 
%
we design a lightweight RNN-based network, because U-Net is computationally expensive with large number of paramaters, making it less suitable for real-time applications where efficiency is the key. 
Our backbone consists of only a stacked GRU\cite{cho2014GRU_ACL} with two cells, followed by a linear layer. 
Despite the simple architecture, as demonstrated in Table \ref{tab:euroc_eval}, our network outperforms U-Net, offering better performance with a significantly smaller architecture.
%
%

%
%

\subsection{Implementation Details}
In practice, we process a window of IMU readings at once, instead of one-by-one, because the network needs context information from IMU readings. However, the window can't be too long either, because the drift inevitably becomes larger as the window is larger, even with correction from predicted bias.  
We choose one-second window as the window size, inspired the choice of \cite{Liu2020RAL} and \cite{Buchanan2023RAL}, striking a balance between capturing sufficient temporal information and maintaining the system online performance. 


For network training, we allow overlapping between IMU windows taken from the full IMU reading sequence, so the network can see more IMU reading patterns. However, too much overlapping provides very similar data, slowing down the training without clear benefit. 
%
%
In practice, we find 50\% overlapping to be a good choice. 
%

The training uses Adam optimizer\cite{kingma14_adam_ICLR}, with learning rate of $3 \times 10^{-5}$, taking 6 hours on an NVIDIA A4500 GPU.
The noise schedule is linearly spaced between $\beta_1 = 0.0001$ and $\beta_T = 0.02$, with the model trained for 1000 steps, following the default setting in \cite{Ho2020NeurIPS}. 

%
%
%

%
For sampling, we select DDIM \cite{Song2020ArXIV} to save sampling steps while maintaining the performance.
We use only 25 sampling steps for bias generation, in contrast to the typical 1000 steps required by standard DDPM sampling \cite{Ho2020NeurIPS}.
%





\subsection{Acquire Bias Ground Truth Data}

To train the model, we require the ground truth bias at the IMU rate. 
Although the bias is not directly measured by the sensor since it is observable \cite{Martinelli2012TRO}, it can be recovered through joint optimization of IMU data and other sensors inputs.
%
Many VINS systems, such as OpenVINS \cite{Geneva2020ICRA}, OKVIS\cite{leutenegger2015_okvis_IJRR} and VINS-Mono\cite{qin2018vinsmon_TRO}, provide reliable bias estimates as part of their state estimation.
When additional sensors like LiDAR \cite{ramezani2020_ncd_IROS} or external motion capture system \cite{Burri2016euroc_IJRR} are available, the bias estimation can be further refined. 
%

Although these bias estimates are typically provided at the frame rate, we can interpolate them to match the higher frequency of the IMU.
Since IMU bias tends to change slowly over time, the interpolated values offer sufficient accuracy for the use as supervision during training.
%

Empirically, we find that bias recovered through joint optimization and then interpolated to the IMU rate is of high quality. 
When the recovered bias is used to correct the IMU data, it results in better motion integration performance compared to bias predicted by deep learning models trained on integrated motion data.
Therefore, the recovered bias can serve as an effective ground truth signal to guide our network towards better performance.

Moreover, we observe that the recovered bias is continuous and changes slowly, consistent with out prior understanding of IMU bias behavior. This further supports the validity of using the recovered bias as the ground truth for training the model.



\section{Experimental Results}




\begin{table*}[t]
\centering
\caption{Motion estimation result for 1-second window on EuRoC dataset (PRMSE / ROE)}
\label{tab:euroc_eval}
\begin{adjustbox}{max width=\linewidth}
\begin{threeparttable}
\setlength{\tabcolsep}{4.5pt}

\begin{tabular}{lcccccc}
\toprule
\textbf{Sequence} & \textbf{AirIMU}     & \textbf{Ours (RNN)} & \textbf{Direct Regression (RNN)} & \textbf{Random Walk} & \textbf{Ours (UNet)} & \textbf{Direct Regression (UNet)} \\ \midrule 
MH02              & {0.0234} / {0.0789} & \gbf{0.0225} / \gbf{0.0604} & 0.0246 / 0.1370                  & 0.0615 / 0.1380      & \gs{0.0227} / \gs{0.0775}  & 0.0343 / 0.1307                   \\[5pt]
MH04              & {0.0415} / {0.0708} & \gs{0.0410} / \gs{0.0636} & 0.0437 / 0.1551                  & 0.0657 / 0.1413      & \gbf{0.0408} / \gbf{0.0593}  & 0.0462 / 0.1233                   \\[5pt]
V103              & {0.0583} / \gs{0.1884}   & \gbf{0.0561} / {0.1931} & 0.0611 / 0.2369                  & 0.0685 / \gbf{0.1762}    & \gs{0.0577} / {0.2185}  & 0.0639 / 0.2574                   \\[5pt]
V202              & 0.0851 / \gs{0.2157}   & \gs{0.0703} / {0.2557} & 0.0777 / 0.3010                  & 0.0813 / \gbf{0.1877}    & \gs{0.0703} / {0.2627}  & \gbf{0.0664} / 0.4179                 \\ \midrule
Average           & {0.0521} / \gbf{0.1385} & \gbf{0.0475} / \gs{0.1432} & 0.0510 / 0.2075                  & 0.0693 / 0.1608      & \gs{0.0479} / {0.1545}  & 0.0527 / 0.2323                   \\ 
\bottomrule
\end{tabular}

\begin{tablenotes} \footnotesize
\item[\gbf{*}] the best performance 
\item[\gs{*}] the second best performance
\item[{*}] V101 not tested as its ground truth accuracy is limited, as reported in \cite{Geneva2020ICRA}
\end{tablenotes}
\end{threeparttable}
\end{adjustbox}
\end{table*}

We conduct our experiments on the EuRoC dataset\cite{Burri2016euroc_IJRR}, using the same training and testing splits as prior studies \cite{Qiu2024Arxiv}.
For evaluation metrics, we use relative Positional RMSE (PRMSE in meters) for position and Relative Orientation Error (ROE in degrees), consistent with established conventions in \cite{Qiu2024Arxiv}.


We compare the performance with the following baselines:
\begin{itemize}
    \item AirIMU\cite{Qiu2024Arxiv}, a recent work that predicts bias through indirect supervision using integrated motion. This method achieves state-of-the-art result on EuRoC dataset, outperforming previous work that also use indirect motion supervision \cite{Brossard2020RAL}\cite{Zhang2021ICRA}, as well as model-free methods \cite{Liu2020RAL}. As expected, the model-free approach, which relies on motion patterns, performs significantly worse on EuRoC dataset. By comparing our method with AirIMU, we indirectly compare it with model-free methods as well.
    
    
    \item Random walk modeling baseline: For this baseline, we use noise density and random walk rate parameters from offline calibrated results provided by the EuRoC dataset.
    Since the random walk model treats bias as a stochastic process, its actual performance is difficult to evaluate directly. 
    We provide a strong baseline as the performance upper bound. We take the ground truth bias at the start of each IMU window, and sample multiple bias changes according to the random walk model. 
    The final bias is the sum of the initial ground truth bias and the sampled bias changes.
    After removing the sampled bias from the IMU readings and integrating the result, we select the best result for each window. 
    It should be noted that this is not a practical algorithm, as it relies on the ground truth data to choose the optimal result.
    In our experiments, we sample bias changes 50 times per window. 
    
    \item Direct bias regression. This method follows similar approach to \cite{Buchanan2023RAL}, where the network directly regresses bias values using the ground truth as supervision. 
    %
    For a fair comparison, we use the same network architecture as our model, with minimal changes to the output layer to match dimensions required for regression.

We do not compare directly with the results from \cite{Buchanan2023RAL} because they only show results using predicted bias in a factor graph optimization framework with visual observations, and their code is not publicly available. 
\end{itemize}

The results are presented in table \ref{tab:euroc_eval}. 
Since our model uses a probabilistic formulation, its predictions are samples from the learned IMU-conditioned bias distribution. Thus, the metric reported in the table is averaged over 50 runs. 
Our model achieves improved performance in terms of position error and the second-best orientation error.
Our result is better than the strong random walk baseline, demonstrating that our bias model is more accurate than commonly used random walk model in its best case. 
Compared with direct regression baseline, our model with almost the same network has better performance. This shows that our probabilistic model formulation can better captures the problem nature than the regression formulation, thus leading to the improved performance.

In comparison to AirIMU, our model has better position error but worse orientation error, resulting in a similar overall performance. As AirIMU is better than the RNN direct regression baseline, who has similar backbone of TCN and GRU, the indirect supervision can offer better accuracy than the direct supervision. However, as we will show in Section \ref{exp:compare_indirect}, indirect supervision method may suffer from spurious prediction. Our method uses the direct supervision while having similar performance to indirect supervision method, combining the best of two methods. This is thanks to our probabilistic formulation implemented with diffusion model, instead of the existing regression formulation. 

\subsection{Diffusion Model vs. Indirect Regression}
\label{exp:compare_indirect}

\begin{figure}[t]
\includegraphics[width=0.5\textwidth]
{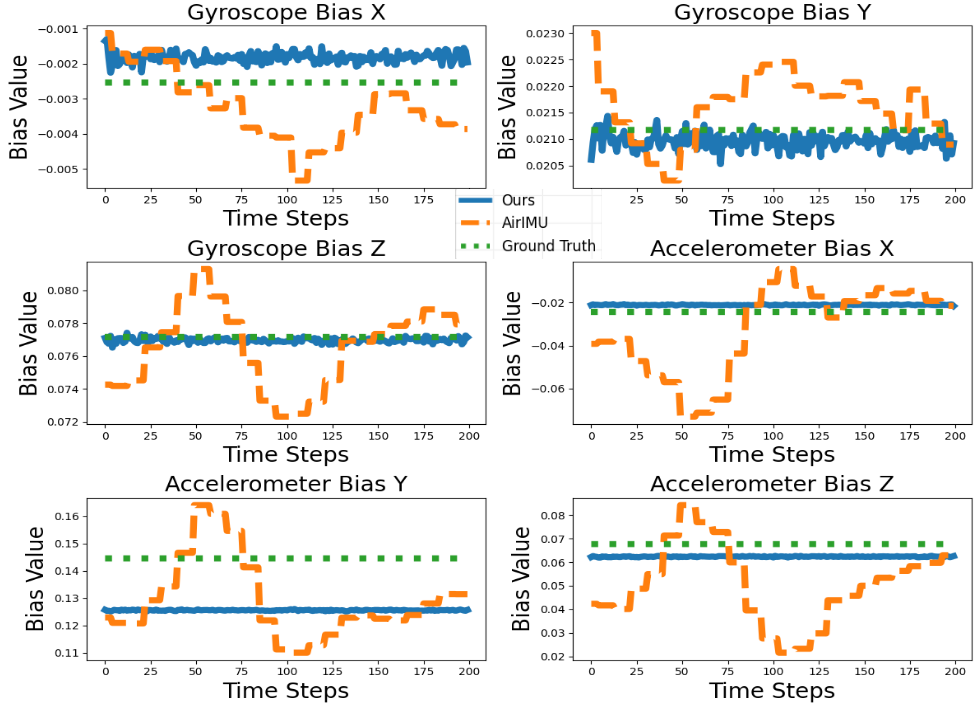}
\caption{Bias prediction result for our model and AirIMU in an one-second window}
\label{fig:res_bias_faithful}
\end{figure}

In this experiment, we compare the bias predictions from our method with those from AirIMU \cite{Qiu2024Arxiv}. 
As mentioned earlier, using indirect supervision through integrated motion can result in spurious bias predictions, as the network may predict unrealistic bias values to optimize motion integration.
While this is not an issue when the end goal is accurate integrated motion, it raises concerns about its generalization ability. 
If the predicted correction does not correspond to the actual IMU bias, it may not capture intrinsic IMU properties and therefore may not generalize well to new data.


We validate this concern in the experiment. As an example, we randomly select one-second IMU reading window and plot the predicted bias values alongside the ground truth. In Fig. \ref{fig:res_bias_faithful}, our prediction match more closely to the ground truth in both magnitude and changing pattern, In contrast, AirIMU's predictions show abrupt changes, violating our prior knowledge of IMU bias.   



%


\begin{table}[h!]
\centering
\begin{tabular}{|c|c|c|}
\hline
\textbf{Model} & \textbf{Parameters (Millions)} & \textbf{Inference Time (ms)} \\ \hline
U-Net Architecture & 42.8 & 170 \\ \hline
RNN Architecture  & 2.2  & 145 \\ \hline
\end{tabular}
\caption{Comparison of model parameters and inference time between U-Net and RNN architectures.}
\label{tab:model_comparison}
\end{table}

\subsection{Timing on Embedded Device}
We evaluate the timing of our pipeline on NVIDIA Jetson AGX Orin embedded device for both the U-Net architecture and the proposed lightweight RNN model. As shown in Table \ref{tab:model_comparison}, the U-Net model has 42.8 million parameters and requires 170 ms for inference, whereas the RNN model is significantly smaller, with 2.2 million parameters, and achieves a faster inference time of 145 ms. This demonstrates the efficiency of the RNN model in terms of both model size and speed.
%


Low-speed real-world applications can benefit from this inference time, such as agriculture and warehouse robots. For
more demanding scenarios such as high-speed drone, further optimization is necessary.
%

\section{Conclusions and Future Work}
In this paper, we have introduced a conditional probability distribution formulation for the IMU bias modeling. Based on this formulation, we have designed a conditional diffusion model to predict the bias from IMU reading, and used it for inertial-only odometry (IOO). Compared with classical random walk bias model and regression based neural network, our model shows better performance and more faithful prediction,
which has been validated on the EuRoC dataset, showing the effectiveness of our probabilistic formulation. 
%
Although we treat the bias as IMU-conditioned probability distribution, there is more work to be done to leverage the probability distribution to make better bias prediction, rather than taking one random sample as the output.
Another direction for future work is to explore how to provide uncertainty for the prediction. 
In all, we believe our new probabilistic formulation for IMU bias modeling opens up new opportunity to capture IMU bias and benefits the field of IOO.

{
\bibliographystyle{packages/IEEEtran}
\bibliography{libraries/related,libraries/rpng}
}

\end{document}